\documentclass{article}
\usepackage{style/spconf,amsmath,graphicx}
\usepackage{multicol}
\usepackage[utf8x]{inputenc} 
\usepackage{amsfonts}
\usepackage{multirow}
\usepackage{makecell}
\usepackage{url}
\usepackage{float}

\def\x{{\mathbf x}}
\def\y{{\mathbf y}}

\title{GaitMixer: Skeleton-based Gait Representation Learning via Wide-spectrum Multi-axial Mixer}
\twoauthors
 {Ekkasit Pinyoanuntapong, Ayman Ali, Pu Wang, Minwoo Lee}
	{University of North Carolina at Charlotte\\
 \{epinyoan, aali26, pu.wang, minwoo.lee\}@uncc.edu}
 {Chen Chen}
	{University of Central Florida \\
 chen.chen@crcv.ucf.edu}
\begin{document}
%
\maketitle
\begin{abstract}
Most existing gait recognition methods are appearance-based, which rely on the silhouettes extracted from the video data of human walking activities. The less-investigated skeleton-based gait recognition methods directly learn the gait dynamics from  2D/3D human skeleton sequences, which are theoretically more robust solutions in the presence of appearance changes caused by clothes, hairstyles, and carrying objects. However, the performance of skeleton-based solutions is still largely behind the appearance-based ones. This paper aims to close such performance gap by proposing a novel network model, GaitMixer, to learn more discriminative gait representation from skeleton sequence data. In particular, GaitMixer follows a heterogeneous multi-axial mixer architecture, which exploits the spatial self-attention mixer followed by the temporal large-kernel convolution mixer to learn rich multi-frequency signals in the gait feature maps.  Experiments on the widely used gait database, CASIA-B, demonstrate that GaitMixer outperforms the previous SOTA skeleton-based methods by a large margin while achieving a competitive performance compared with the representative appearance-based solutions.  Code will be available at \url{https://github.com/exitudio/gaitmixer}

\end{abstract}
\section{Introduction}
Unlike short-distance biometrics (e.g., fingerprints, facial, iris, palm, and finger vein patterns), gait can be recognized from a distance without the subject's cooperation or interference.
Such long-distance biometrics has a huge potential to extend its applications to forensic identification, access control, and social security. The gait recognition methods are generally  either appearance-based or skeleton-based. Appearance-based approaches \cite{GaitNet}\cite{GaitSet}\cite{GaitPart}\cite{cross_view_cnn_silhouette} utilize background subtraction to obtain silhouettes from a video sequence, which are further analyzed using carefully-designed network models for gait representation learning. 
On the other hand, skeleton-based approaches \cite{PTSN}\cite{GaitGraph}\cite{GaitGraph2} utilize the skeleton sequences extracted from 2D/3D pose estimators as the inputs to learn effective gait representations. Theoretically,  skeleton-based methods are more robust to appearance variations caused by hairstyles, carrying objects, and clothes. However,  the skeleton-based approaches, which still do not receive sufficient attention, yield a large performance gap compared with the appearance-based counterparts. 


\begin{figure}[t]
\centerline{\includegraphics[width=.4\textwidth]{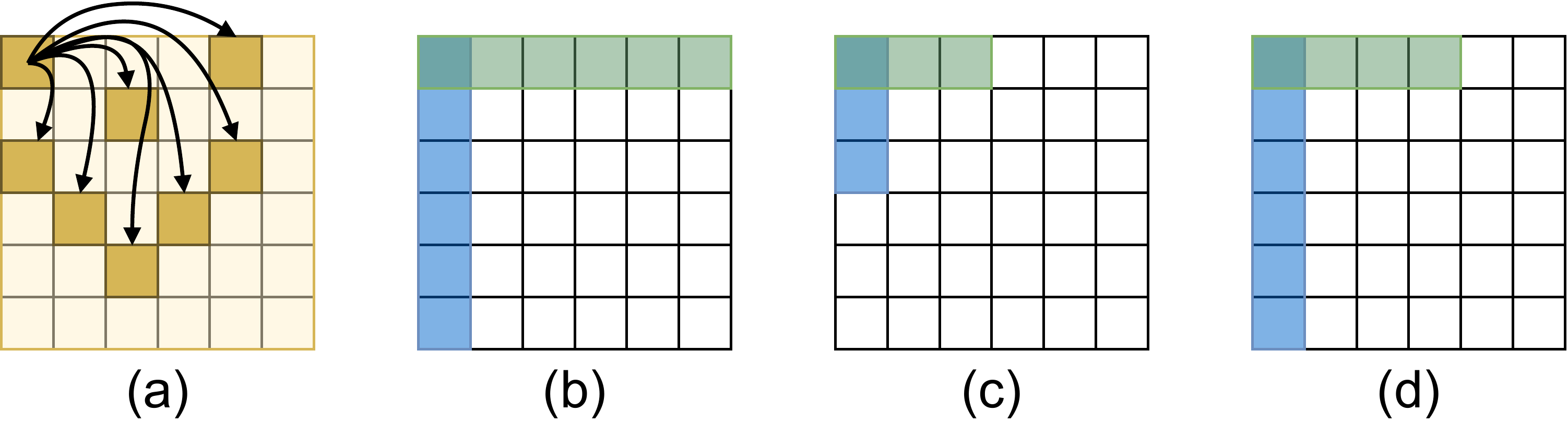}}
\caption{(a) Global self-attention token mixing \cite{ViT}. (b) Self-attention mixing along H and W axes \cite{ViViT}. (c) Convolution-mixing along H and W axes \cite{GaitGraph} (d) Heterogeneous multi-axial mixer (ours). }
\label{fig:mixer}
\end{figure}


\begin{figure*}[ht]
 \centering
  \includegraphics[width=0.8\textwidth]{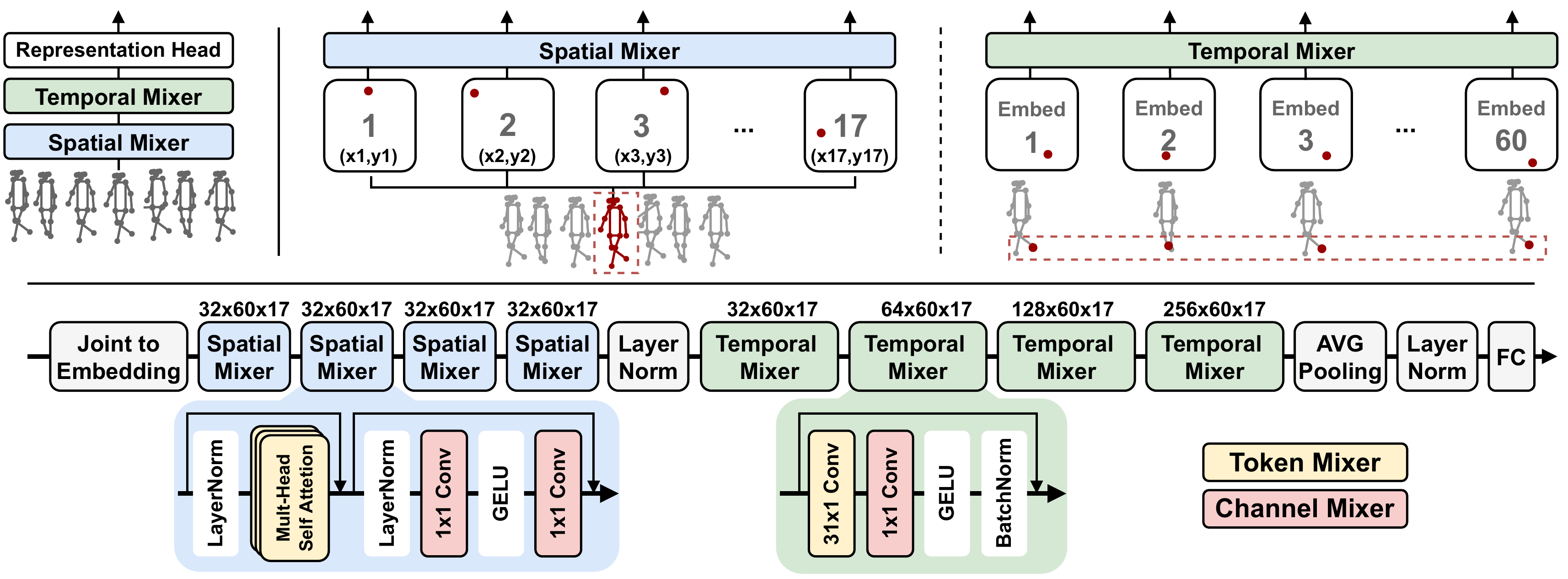}
  \caption{(top) GaitMixer consists of a spatial mixer followed by a temporal mixer. (bottom) The detailed network architecture of the spatial self-attention mixer and large-kernel convolution mixer}
  \label{fig:architecture}
\end{figure*}

To close this gap, this paper tries to exploit more effective gait feature encoders by proposing the multi-axial mixer, which is a generic transformer-like architecture that mixes the feature patches (i.e., tokens) along each axis of the feature space, respectively, e.g., width-wise, height-wise, and channel-wise axes in image feature space. Many recent high-performance network backbones can be considered as the special cases of multi-axial mixer based on what types of mixing functions are applied, which mainly include convolution \cite{GaitGraph}\cite{st-gcn}\cite{ResGCN}, self-attention \cite{st-tr}\cite{PoseFormer}, and multi-layer perceptrons (MLP) \cite{mlp-mixer} \cite{MorphMLP} as shown in Fig. \ref{fig:mixer}. Multi-axial mixers have been demonstrated to achieve SOTA performance in image classification and video recognition tasks, while significantly reducing computation complexities compared with other competitive network models, such as vision transformers. Despite their promising features, current multi-axial mixers generally exploit the homogeneous architecture design, where the same type of mixing functions (e.g., either convolution, self-attention, or MLP) is applied along each feature space axis. Such design, however, has limited capacity to learn multi-frequency features. In particular, it has been established that convolutions focus more on local information and therefore are good learners for high-frequency features \cite{ScaleUp31x31}.  Self-attentions, on the contrary, are designed to model long-range interactions and are more capable to capture low-frequency signals (global information) in feature map \cite{HiLoAttention}.

In this paper, we propose GaitMixer, a novel \emph{heterogeneous} spatial-temporal axial mixer, which can effectively learn the discriminative gait representation by capturing both high-frequency and low-frequency features.  In particular, GaitMixer consists of a spatial self-attention mixer and a temporal large-kernel axial mixer (Fig.~\ref{fig:architecture}). The spatial axial mixer learns interactions among the joints within each skeleton frame. The temporal axial mixer models the interactions among the temporal tokens of each single joint at different time indices. Experiments on the widely used gait database, CASIA-B, demonstrate that GaitMixer 
outperforms the previous SOTA skeleton-based \cite{GaitGraph} methods by 12\% on average, while achieving a competitive performance compared with the representative appearance-based solutions.

\section{RELATED WORK}
\label{sec:related_work}


Appearance-based approaches extract binary images of a human silhouette from the source images by subtracting static background \cite{Silhouette_analysis_based}. 
GaitNet \cite{GaitNet} integrates silhouette extraction into the model as an end-to-end network for gait recognition. GaitSet \cite{GaitSet} decouples the temporal continuous sequence by learning identity information from the set of independent frames to be immune to permutation of frames and be able to integrate frames from different videos. While the majority of methods \cite{GaitNet}\cite{GaitSet}\cite{cross_view_cnn_silhouette} take the entire shape as input, more recent approaches GaitPart \cite{GaitPart} focus on each part of the body individually assuming that each part of human body needs its own spatial-temporal learning by separating silhouette into several parts horizontally. Skeleton-based approaches (i.e., model-based approaches) use skeleton data as the model inputs. In the early work, pose-based temporal-spatial network (PTSN) \cite{PTSN} utilizes a long-short term memory (LSTM) to capture the dynamic information and CNN to learn static information of a gait sequence in parallel. PoseGait \cite{PoseGait} utilizes 3D pose estimated from images in order to be invariant to view changes, along with hand-crafted features including joint angle, limb length, and joint motion. The most recent methods, GaitGraph \cite{GaitGraph} and GaitGraph2 \cite{GaitGraph2}, adopt graph convolution neural networks (GCNs) for gait recognition, inspired by the successes of GCNs in action recognition tasks.

\section{GaitMixer: Heterogeneous Wide-Spectrum Spatial-Temporal Mixer}
\label{sec:method}

To effectively learn both high-frequency and low-frequency gait features, we introduce the GaitMixer, a heterogeneous spatial-temporal axial mixer architecture. As shown in Fig.  \ref{fig:architecture}, GaitMixer consists of a spatial self-attention mixer and a temporal large-kernel axial mixer (Fig.~\ref{fig:architecture}). The spatial axial mixer only learns interactions among the joints within each skeleton frame. The $d_y$-dimensional spatial representation $y^{t} \in \mathbb{R}^{|J| \times d_y}$ for each skeleton frame $t$ with $|J|$ joints is learned after $B_S$ self-attention blocks. Then, the representations of $T$ skeleton frames within a gait sequence are concatenated into $z \in \mathbb{R}^{|J| \times T \times d_y}$, which is then forwarded to a temporal axial mixer to capture the interactions among the tokens of each single joint at different temporal indices. The temporal axial mixer consists of $B_T$ one-dimensional large-kernel convolution blocks. Moreover, to simplify our GaitMixer architecture, both spatial and temporal mixers adopt the isotropic design, which does not perform feature downsampling and maintains the same feature resolutions at all layers.

\subsection{Spatial Mixer with Axial Self-attention}


The spatial mixer module aims to learn a high dimensional representation embedding
from each skeleton frame. Although self-attention tends to capture low-frequency (or global) features, our experiments demonstrate that self-attention is sufficient to learn both high-frequency and low-frequency signals in the feature map along spatial axis. This also indicates that self-attention can effectively model both short-range and long-range inter-joint dependencies. Given a 2D skeleton with joints \(J\), we consider each joint (\emph{i.e}., \(x\) and \(y\) coordinates) as a spatial token (with 2 channels) and perform the feature extraction among all \(|J|\) spatial tokens by following the isotropic transformer pipeline. Specifically, the spatial taken $x_i \in \mathbb{R}^2$  is passed through a trainable linear projection, which maps each token to a high dimension embedding $\x_i \in \mathbb{R}^{d_x}$. Then, the spatial token embeddings of each skeleton frame $\x=(\x_1, \x_2,\dots,\x_{|J|})$ are mixed by inter-token dot product attentions to generate an output sequence $\y=(\y_1,\y_2,\dots,\y_{|j|})$ where $\y_i \in \mathbb{R}^{d_y}$. Running $h$ self-attentions in parallel leads to the multihead self-attention with $h$ heads, where the outputs of the attention heads are concatenated and projected into the expected dimensions.


\subsection{Temporal Mixer with Large-kernel Convolution} 
The essence of a walking sequence is composed of multiple short repeated cycles. In the temporal axis, self-attention may not be able to capture wide-band multi-frequency features, considering that the global receptive field of self-attention is much easier to capture low-frequency features. It demands a large amount of data for self-attention to establish the desirable locality inductive bias that is the key to learn high-frequency features. To learn both high-frequency and low-frequency temporal data, we utilize large kernel depth-wise separable convolution in the temporal mixer as illustrated in Fig.~\ref{fig:architecture}. In general, convolution neural networks tend to capture high-frequency (local) features, however, the large kernel allows the model to also learn low-frequency features. 60 frames are used in the temporal model which covers around 4 cycles of walking. A large one-dimension kernel with size of $31 \times 1$ is used to capture mid-range information (around 2 walking cycles). A reverse padding with size of 30 is applied to keep the temporal dimension the same. The temporal mixer only communicates with all frames along the temporal axis of the same joints. A temporal mixer is composed of two types of axial mixers. First, a token mixer is implemented by a depth-wise convolution that learns all embeddings only in the same channel. Next, a channel mixer is a $1\times1$ convolution that learns only specific embedding along all channels.


\subsection{Representation head and Loss Function} 
We apply average pooling along spatial and temporal dimensions to reduce the output from $\x_{temp} \in \mathbb{R}^{F\cdot (J\times c)}$ to $\x_{hidden} \in \mathbb{R}^{c}$, where $F$ and $J$ are number of frames and joints respectively. The number of channels $c$ is set to 256. Finally, layer norm, fully connected layer, and \(l^2\)-norm are applied respectively and return the feature embedding in 128 dimensions. To learn the discriminative gait representation, we apply the triplet loss with multi-similarity miner.

\begin{table*}[h]
\caption{Averaged Rank-1 accuracies on CASIA-B per probe angle excluding identical-view cases.}
\centering
\scalebox{0.7}{
\begin{tabular}{c|c|c|c|c|c|c|c|c|c|c|c|c|c} 
\hline
\multicolumn{2}{c|}{\textbf{Gallery NM\#1-4}}                & \multicolumn{11}{c|}{\textbf{0°-180°}}                                                                                                                                        & \textbf{mean}  \\ 
\hline
\multicolumn{2}{c|}{\textbf{Probe}}                          & \textbf{0°}   & \textbf{18°}  & \textbf{36°}  & \textbf{54°}  & \textbf{72°}  & \textbf{90°}  & \textbf{108°} & \textbf{126°} & \textbf{144°} & \textbf{162°} & \textbf{180°} &                \\ 
\hline
\multirow{4}{*}{\textbf{NM\#5-6}} & PoseGait \cite{PoseGait}                 & 55.3          & 69.6          & 73.9          & 75.0          & 68.0          & 68.2          & 71.1          & 72.9          & 76.1          & 70.4          & 55.4          & 68.7           \\
                                  & GaitGraph  \cite{GaitGraph}              & 85.3          & 88.5          & 91.0          & 92.5          & 87.2          & 86.5          & 88.4          & 89.2          & 87.9          & 85.9          & 81.9          & 87.7           \\
                                  & GaitGraph2  \cite{GaitGraph2}             & 78.5          & 82.9          & 85.8          & 85.6          & 83.1          & 81.5          & 84.3          & 83.2          & 84.2          & 81.6          & 71.8          & 82.0                                \\
                                  & GaitFormer (ours) & 90.9          & 91.2          & 93.7          & 91.9          & 91.9          & 92.7          & 93.3          & 91.8          & 92.5          & 90.5          & 85.5          & 91.5           \\ 
\cline{2-14}
                                  & \textbf{GaitMixer(ours)} & \textbf{94.4} & \textbf{94.9} & \textbf{94.6} & \textbf{96.3} & \textbf{95.3} & \textbf{96.3} & \textbf{95.3} & \textbf{94.7} & \textbf{95.3} & \textbf{94.7} & \textbf{92.2} & \textbf{94.9}  \\ 
\hline
\multirow{4}{*}{\textbf{BG\#1-2}} & PoseGait   \cite{PoseGait}              & 35.3          & 47.2          & 52.4          & 46.9          & 45.5          & 43.9          & 46.1          & 48.1          & 49.4          & 43.6          & 31.1          & 44.5           \\
                                  & GaitGraph    \cite{GaitGraph}            & 75.8          & 76.7          & 75.9          & 76.1          & 71.4          & 73.9          & 78.0          & 74.7          & 75.4          & 75.4          & 69.2          & 74.8           \\
                                  & GaitGraph2 \cite{GaitGraph2}              & 69.9          & 75.9          & 78.1          & 79.3          & 71.4          & 71.7          & 74.3          & 76.2          & 73.2          & 73.4          & 61.7          & 73.2                                \\
                                  & GaitFormer (ours) & 82.5          & 83.2          & 85.7          & 85.7          & 84.2          & 80.2          & 78.9          & 82.6          & 82.2          & 78.6          & 71.3          & 81.4           \\ 
\cline{2-14}
                                  & \textbf{GaitMixer(ours)} & \textbf{83.5} & \textbf{85.6} & \textbf{88.1} & \textbf{89.7} & \textbf{85.2} & \textbf{87.4} & \textbf{84.0}          & \textbf{84.7} & \textbf{84.6} & \textbf{87.0} & \textbf{81.4} & \textbf{85.6}  \\ 
\hline
\multirow{4}{*}{\textbf{CL\#1-2}} & PoseGait  \cite{PoseGait}               & 24.3          & 29.7          & 41.3          & 38.8          & 38.2          & 38.5          & 41.6          & 44.9          & 42.2          & 33.4          & 22.5          & 36.0           \\
                                  & GaitGraph \cite{GaitGraph}               & 69.6          & 66.1          & 68.8          & 67.2          & 64.5          & 62.0          & 69.5          & 65.6          & 65.7          & 66.1          & 64.3          & 66.3           \\
                                  & GaitGraph2  \cite{GaitGraph2}             & 57.1          & 61.1          & 68.9          & 66            & 67.8          & 65.4          & 68.1          & 67.2          & 63.7          & 63.6          & 50.4          & 63.6                                \\
                                  & GaitFormer (ours) & 76.1          & 80.3          & 81.0          & 78.2          & 77.7          & 76.6          & 77.4          & 75.8          & 76.5          & 75.7          & 77.2          & 77.2           \\ 
\cline{2-14}
                                  & \textbf{GaitMixer(ours)} & \textbf{81.2} & \textbf{83.6} & \textbf{82.3} & \textbf{83.5} & \textbf{84.5} & \textbf{84.8} & \textbf{86.9} & \textbf{88.9} & \textbf{87.0} & \textbf{85.7} & \textbf{81.6} & \textbf{84.5}  \\
\hline
\end{tabular}
}
\label{tab:skeleton-based}
\end{table*}

\begin{table}
\vspace{-0.4cm}
\caption{Averaged Rank-1 accuracies on CASIA-B comparison with both appearance-based and skeleton-based
methods}
\centering
  \resizebox{0.8\linewidth}{!}
  {
  \begin{tabular}{c|c|c|c|c|c}
\hline
\multirow{2}{*}{}              & \multirow{2}{*}{\textbf{Method }} & \multicolumn{4}{c}{\textbf{Probe }}                             \\ 
\cline{3-6}
                                             &                                   & \textbf{NM}   & \textbf{BG}   & \textbf{CL}   & \textbf{Mean}   \\ 
\hline
\multirow{3}{*}{\textbf{\makecell{Appearance \\based}}} & GaitNet   \cite{GaitNet}                        & 91.6          & 85.7          & 58.9          & 78.7           \\
                                             & GaitSet   \cite{GaitSet}                        & 95.0            & 87.2          & 70.4          & 84.2            \\
                                             & GaitPart  \cite{GaitPart}                        & \textbf{96.2}          & \textbf{91.5}          & 78.7          & \textbf{88.8}            \\ 
\hline
\multirow{4}{*}{\textbf{\makecell{Skeleton \\based}}}    & PoseGait                          & 68.7          & 44.5          & 36.0          & 49.7           \\
                                             & GaitGraph                         & 87.7          & 74.8          & 66.3          & 76.3           \\
                                             & GaitGraph2                        & 82.0          & 73.2          & 63.6          & 72.9          \\
                                             & \textbf{GaitFormer (ours)} & 91.5          & 81.4          & 77.2          & 83.4           \\ 
\cline{2-6}
                                             & \textbf{GaitMixer (ours)}         & 94.9 & 85.6 & \textbf{84.5} & \textbf{88.3}  \\
\hline
\end{tabular}
}
\label{tab:appearance-based}
\end{table}
\begin{figure}[t]
  \includegraphics[width=8.8cm]{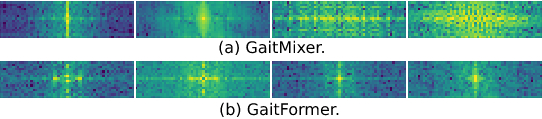}
  \caption{2D FFT of GaitMixer and GaitFormer feature maps of 4 channels. Higher temperature indicates larger magnitude. Pixels closer to the center represent lower frequencies}
\label{fig:fft}
\end{figure}
\section{EXPERIMENTS}
\subsection{Dataset}
\textbf{CASIA-B} \cite{CASIA-B} has been widely adopted as a multi-view, RGB, and silhouette gait dataset. The data acquisition is performed by 124 individuals from 11 viewing angles ranging from 0 to 180 with 18 angle differences. To mimic typical daily walking conditions, each subject performs six sequences of normal walking (NM), two sequences of walking with a coat (CL), and two sequences of walking with carrying a bag (BG). For each individual, ten sequences are captured from each view angle. This paper follows a widely-used test protocol \cite{GaitNet}\cite{GaitSet}\cite{GaitPart}\cite{GaitGraph}\cite{GaitGraph2}\cite{PoseGait}, which uses the data of the first 74 subjects' sequences for the training and the remaining 50 subjects' sequences for testing. Furthermore, the test dataset is divided into gallery and probe sets. The gallery set includes the first four sequences of the normal walking condition. The probe set consists of the last two sequences of normal walking, two walking with a coat on, and walking with carrying a bag. Finally, the results are reported for all viewing angles. 

\subsection{Implementation Details}

\noindent\textbf{Training Details} HRNet \cite{hrnet} is used as a 2D human pose estimator. We follow data augmentations from GaitGraph\cite{GaitGraph} and add normalization of the joint position in $(x, y)$-coordinates by dividing 320 which is the width of the original videos to  input data while keeping the aspect ratio. Adam optimizer is used with $6\mathrm{e}{-3}$ learning rate with 1-cycle learning rate and $1\mathrm{e}{-5}$ weight decay. We are using a balanced batch sampler to sample the number of walking data per person equally. The batch size is (74, 4), denoting 74 people and 4 walking samples per person. \noindent\textbf{Testing}. 
Each gait testing sample contains 60 frames selected from the middle of the sequence data. The test set is separated into probe and gallery.  Both are fed into the model to obtain the feature representations. The ID of the gallery representation that has the smallest cosine distance from the probe will be the predicted ID of the probe. 


\subsection{Comparison with the SOTA Methods}
To demonstrate the superior performance of GaitMixer as a heterogeneous multi-axial mixer model, we also build GaitFormer, which is a homogeneous multi-axial mixer that adopts self-attention for both spatial and temporal axes.
In Fig. \ref{fig:fft}, we visualize the frequency magnitude of the output feature maps from GaitMixer and GaitFormer, respectively. It can be observed that GaitMixer concentrates on both high-frequency and low-frequency components along both temporal and spatial axes in feature maps. This confirms the superior capacity of GaitMixer to capture features in wide-spectrum bands. GaitFormer, however, cannot effectively model the high-frequency feature components. The performance comparisons between our approaches and the SOTA skeleton-based methods are shown in Table \ref{tab:skeleton-based}. It is shown that our multi-axial mixer models outperform the existing solutions by a large margin in both cross-view and cross-walking-condition cases. Moreover, GaitMixer achieves better recognition accuracy than GaitFormer because GaitMixer can jointly exploit the heterogeneous mixing at different feature space dimensions. Table \ref{tab:appearance-based} shows a competitive performance of our skeleton-based methods, compared with the representative appearance-based methods. GaitMixer achieves much higher accuracy than all appearance-based approaches in wearing coat condition. It is due to the inherent robustness of skeleton data against large appearance changes.



\begin{figure}[t]
\begin{minipage}[b]{.47\linewidth}
  \centering
  \centerline{\includegraphics[width=4.3cm]{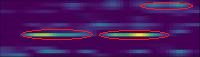}}
\end{minipage}
\hfill
\begin{minipage}[b]{0.50\linewidth}
  \centering
  \centerline{\includegraphics[width=4.4cm]{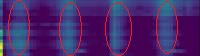}}
\end{minipage}
\begin{minipage}[b]{1\linewidth}
  \centering
  \centerline{\includegraphics[width=8.8cm]{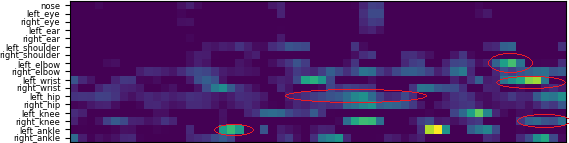}}
\end{minipage}
\caption{\textbf{Grad-CAM} \cite{Grad-CAM} visualizations. \textbf{Top-left}: GaitGraph \cite{GaitGraph}. \textbf{Top-right}: GaitFormer. \textbf{Bottom}: GaitMixer. X-axis represents frames 1 to 60 and Y-axis represents 17 joints. Features with higher contributions have higher heat temperatures.}
\label{fig:gradcam}
\end{figure}

\subsection{Visualization}
We use class activation map (Grad-CAM) \cite{Grad-CAM} to show which parts of the input gait sequence contribute most to the final recognition result. As shown in Fig.~\ref{fig:gradcam} (bottom),  GaitMixer focuses on continuous joint sequences with a variety of different temporal windows, thus capturing short-, mid, and- long-range temporal feature interactions. Moreover, GaitMixer also pays attentions to a diverse set of joints except ears, eyes, and nose, which is also as expected because the landmarks on face are not relevant to the gait dynamics. As shown in  Fig.~\ref{fig:gradcam} (top-left),  GaitGraph tends to focus on some specific joints over a large temporal window and it also exploits the features from face landmarks for gait recognition. Both limitations could degrade the performance of GaitGraph. GaitFormer (Fig.~\ref{fig:gradcam} (top-right)) pays more attention to certain skeleton frameworks without capturing rich spatial-temporal feature interactions. This can be the key contributing factor that affects its performance.


\section{Conclusion}
In this paper, we present GaitMixer model, a novel heterogeneous multi-axial architecture combining a spatial self-attention mixer and a large kernel temporal convolution mixer to capture both high-frequency and low-frequency dynamics of gait data. Our approach achieves the best accuracy on the well-known CASIA-B gait dataset for all conditions when compared to previous skeleton-based methods and is superior to appearance-based approaches with coats conditions.

\bibliographystyle{style/IEEEbib}
\bibliography{ref/skeleton-based,ref/appearance-based,ref/mixer,ref/action-recognition_3dLift,ref/dataset_tool}
\end{document}